\pdfoutput=1
\documentclass[10pt,a4paper]{article}
\usepackage{amsfonts}
\usepackage{amssymb}
\usepackage{graphicx}
\usepackage{cite}
\usepackage{authblk}
\usepackage{tikz}

\usepackage[english]{babel}
\usepackage{amsmath}
\usepackage{amsfonts}
\usepackage{amssymb}
\usepackage{mathrsfs}
\usepackage{makeidx}
\usepackage{graphicx}
\usepackage[left=2cm,right=2cm,top=2cm,bottom=2cm]{geometry}
\usepackage{bm} 
\usepackage{color}
\usepackage{subfigure}
\usepackage{tikz}
\usepackage{algorithmic}
\usepackage{algorithm}
\usepackage[justification=centering]{caption}
\usepackage{psfrag}
\usepackage{algorithm, algorithmic}
\usepackage{multirow}

\providecommand{\ve}[1]{{\bm {#1}}} %
\providecommand{\mat}[1]{{\bm {#1}}} %

\providecommand{\est}[1]{{\widehat{#1}}}

\DeclareMathOperator*{\argmax}{arg\,max}

\newcommand{\Real}{\mathbb{R}}

\newcommand{\T}{\textsf{T}}

\title{Post-hoc labeling of arbitrary EEG recordings for data-efficient evaluation of neural decoding methods} 

\author{Sebastian Casta\~no-Candamil\footnote{Corresponding authors: sebastian.castano@bltb.uni-freiburg.de, michael.tangermann@blbt.uni-freiburg.de} }
\author{Andreas Meinel}
\author{Michael Tangermann$^*$}
\affil{Brain~State~Decoding~Lab, Department~of~Computer~Science, BrainLinks-BrainTools~Cluster~of~Excellence, University~of~Freiburg, Germany}

\begin{document}
\maketitle

\begin{abstract}
Many cognitive, sensory and motor processes have correlates in oscillatory neural sources, which are embedded as a subspace into the recorded brain signals. Decoding such processes from noisy magnetoencephalogram/electroencephalogram (M/EEG) signals usually requires the use of data-driven analysis methods. The objective evaluation of such decoding algorithms on experimental raw signals, however, is a challenge: the amount of available M/EEG data typically is limited, labels can be unreliable, and raw signals often are contaminated with artifacts. The latter is specifically problematic, if the artifacts stem from behavioral confounds of the oscillatory neural processes of interest.
 
To overcome some of these problems, simulation frameworks have been introduced for benchmarking decoding methods. Generating artificial brain signals, however, most simulation frameworks make strong and partially unrealistic assumptions about brain activity, which limits the generalization of obtained results to real-world conditions. 

In the present contribution, we thrive to remove many shortcomings of current simulation frameworks and propose a versatile alternative, that allows for objective evaluation and benchmarking of novel data-driven decoding methods for neural signals. Its central idea is to utilize post-hoc labelings of arbitrary M/EEG recordings. This strategy makes it paradigm-agnostic and allows to generate comparatively large datasets with noiseless labels. Source code and data of the novel simulation approach are made available for facilitating its adoption.
\end{abstract}


\section{Introduction}
Brain oscillatory phenomena measured with non-invasive imaging techniques, as magneto- or electroencephalography (M/EEG), contain information about underlying neural processes relevant to the neuroscience community~\cite{MeiCasReiTan16,TanReiMei15}. Since the rise of brain-computer interfaces (BCIs), great effort has been put into developing novel techniques for decoding --- typically non-stationary --- neural sources from noisy recordings using linear and nonlinear methods, both for classification and regression tasks. Among them, linear subspace decomposition methods are commonly found. Besides extracting label-informative oscillatory components, these methods often act as a dimensionality reduction step.

The pioneering work on joint covariance diagonalization presented by Fukunaga~\cite{fukunaga2013introduction} and reformulated by de~Cheveign{\'e} and Parra~\cite{de2014joint} serves as a generalized foundation for popular supervised subspace decomposition algorithms utilized in neurosciences, such as the common spatial patterns (CSP) algorithm used for classification tasks\cite{koles1990spatial,lemm:2005}. The relevance of the CSP is not only indicated by its extensive use~\cite{tangermann2012review}, but also by the plethora of derivatives that have been introduced after its original presentation: finite impulse response CSP (FIR-CSP)~\cite{higashi:2013}, sub-band CSP (SBCSP)~\cite{novi:2007}, filter-bank CSP (FBCSP)~\cite{ang:2008}, spectrally weighted-CSP (SPEC-CSP)~\cite{tomioka:2006}, among others. 
On the other hand, the source-power comodulation algorithm SPoC~\cite{SPoCdaehne:2014}, together with its extensions canonical SPoC (cSPoC)\cite{cSPoCdaehne:2014} and multimodal SPoC (mSPoC)~\cite{mSPoCdaehne:2014}, are widely used approaches in the realm of supervised linear decoding methods for regression tasks~\cite{MeiCasReiTan16}. Unsupervised linear neural decoding methods are also extremely popular. Among them, the most widely used may be independent component analysis (ICA) as an approach to blind-source separation~\cite{makeig1996independent}. Specially interesting in the context of the extraction of oscillatory components is spatio-spectral decomposition (SSD) introduced by Haufe et al.~\cite{haufe2014dimensionality}.

Furthermore, nonlinear decoding methods have also been introduced. For example, the work on convolutional neural networks presented by Schirrmeister et al.~\cite{schirrmeister:2017} for classification of motor-imagery tasks and by Lawhern et al.~\cite{lawhern2016eegnet} for classification of visual-evoked potentials, error-related negativity responses, movement-related cortical potentials, and sensory motor rhythms. Further decoding approaches making use of novel machine learning solutions are profusely found in the literature (see review provided in~\cite{schirrmeister:2017}).

\subsection*{Benchmarking and validation of data-driven neural decoding methods}
For the development, validation, and benchmarking of such neural decoding methods, it is desirable to have multichannel datasets with large amounts of labeled data. 

In the literature, two types of testing frameworks prevale. First, frameworks making use of real M/EEG recordings acquired during experimental sessions, and second, those using synthetically generated pseudo-M/EEG signals. Each comes with advantages and deficiencies, as explained below.

\subsubsection*{Real M/EEG recordings} 
Using real M/EEG data has the great advantage that its dynamics, signal-to-noise ratio between oscillatory sources of interest and background activity as well as its non-stationary behavior over time are real. 

However, the amount of real M/EEG data acquired in a single experimental session of a subject maximally lasts a couple of hours. This limited dataset size is rendered even smaller in subsequent processing steps, i.e.,~data segmentation, removal of inter-trial pauses and rejection of artifactual segments. Bigger datasets may be obtained by applying transfer learning techniques, with the aim of merging inter-subject and inter-session data. However, this comes with its own substantial challenges and still is subject to active research~\cite{jayaram2016transfer}. Overall, the relative small dataset size is a clear drawback of using real M/EEG data for the benchmarking of algorithms.

If M/EEG recordings are governed by a varying but known experimental parameter --- such as the intensity of an external stimulus~\cite{SPoCdaehne:2014} --- this parameter can be used as a target variable $\ve{z}$, which serves as epoch-wise labels for decoding correlated oscillatory M/EEG activity. 
Unfortunately, the situation is more difficult, if an M/EEG correlate of an imagery task or open behavior shall be decoded, as these usually lack a \textit{direct} behavioral surrogate. Instead, they may in the best case deliver only a noisy estimate thereof. 
This label noise can have many different origins: subjects may not perfectly follow the experimental instructions, may change their mental strategies to solve a problem, or display varying levels of engagement over time. 
Compared to clean labels $\ve{z}$, the noise contained in label information is known to decrease the performance of decoding algorithms~\cite{CasMeiDaeTan15}. A number of decoding tasks like motor imagery experiments~\cite{hohne2014motor} in BCI, motor performance prediction~\cite{MeiCasReiTan16} or attention decoding~\cite{martel2014} currently are very challenging, and label noise may be part of the problem.
As the experimenter typically neither knows the level of label noise contained in $\ve{z}$ nor can control it, behavioral experiments deliver suboptimal data for the benchmarking of algorithms.

Last but not least, the use of real M/EEG data for benchmarking comes with the drawback that switching between decoding approaches, e.g.,~classification and regression, may require to re-run M/EEG recordings to collect the necessary novel label types.


\subsubsection*{Synthetic pseudo-M/EEG signals}
To avoid the shortcomings of real data, synthetically generated pseudo-M/EEG signals may be also used, as performed in the fields of brain mapping and connectivity analysis~\cite{CasHoeMarAnCasHau15, haufe2016simulation}. Here, the assumption of a linear mapping from the neural source space to the M/EEG sensor space allows to simulate the activity of a neural target source, whose activity overlaps with measurement noise and task-irrelevant brain activity termed \textit{background sources}. Special attention has been dedicated into modeling sources such that they match naturally occurring frequency spectra, e.g.,~to have backround sources reproducing a $1/f$ frequency spectrum and a narrow-band oscillating target source. These semi-realistic simulations need to make strong assumptions about brain dynamics by fixing, e.g., the power ratio between target- and background sources, the noise level on the sensor space, and time series of the sources.  Synthetic datasets typically disregard non-stationarity, which are present in real datasets and which pose substantial challenges for decoding methods. While being sufficient for proof-of-concept purposes~\cite{SPoCdaehne:2014}, these purely synthetic datasets lack the sufficient realism for enabling the analysis of novel decoding algorithms in a real-world scenario.

Simulation strategies for generation of M/EEG time-series have been used extensively in the field of computational neuroscience. Here, physiologically motivated linear and nonlinear stochastic models have been introduced, modeling, e.g.,~dynamics of Alzheimer's disease, epilepsy, or sleeping disorders~\cite{kim2007compact, robinson2002dynamics}. Likewise, data-driven methods for the generation of pseudo-M/EEG time-series have also been presented~\cite{forney2015est}. However, in these models, the notion of spatiality --- provided by real M/EEG --- is typically not utilized. 
Elaborating on this idea, it may be argued that the generated synthetic data may be well-suited to simulate time series activity of target neural sources within a linear projection approach (see the previous paragraph on brain mapping literature), but even then we remain with the problem of selecting realistic properties for background sources and noise.


\subsection*{Post-hoc labeling of paradigm-agnostic M/EEG recordings}
Motivated by the shortcomings of using real M/EEG recordings (few data and noisy labels) as well as of synthetically generated datasets (arbitrary assumptions about neural dynamics), we propose a novel labeled dataset generation framework. It is based on post-hoc labeling of pre-recorded real M/EEG signals and generates novel labels using unsupervised subspace projection methods. As the labels are generated anew, the framework is agnostic wrt.~the original paradigm, under which the M/EEG signals had been recorded, and to its original trial structure.

As a result, our framework offers a highly efficient usage of data (thus yielding potentially larger datasets), completely noiseless labels and real M/EEG dynamics. 
To facilitate the adoption of the framework, source code and datasets are made publicly available.

\section{Methods}
\subsection{Generative model of brain activity}
Neural activity recorded by M/EEG can be represented by means of a linear forward model~\cite{baillet2001electromagnetic,grech2008review}:
\begin{align}\label{eq:genModel}
\mat{X}=\mat{A}\,\mat{S} + \mat{E} \;,
\end{align}

where $\mat{X}\negmedspace\in \,\mathcal{C} \subseteq \Real^{N_c \times N_t}$ is a multivariate signal in the channel space $\mathcal{C}$ describing M/EEG data measured by  $N_c$ M/EEG channels at $N_t$ discrete time samples, $\mat{S} \negmedspace\in\,\mathcal{S} \subseteq \Real^{N_s \times N_t}$ describes the time course of $N_s$ neural sources in the source space $\mathcal{S}$ with covariance matrix  $ \mat{Q} \negmedspace\in\negmedspace \Real^{N_s\times N_s}$, and matrix $\mat{A}\negmedspace\in\negmedspace \Real^{N_c \times N_s}$ describes the linear projection $\mathcal{S} \rightarrow  \mathcal{C}$ of the sources into the sensor space, where the columns of $\mat{A}$, $\ve{a} \negmedspace\in\negmedspace \Real^{N_c}$, are referred to as \emph{spatial patterns}. Furthermore, the matrix $\mat{E}$ contains i.i.d.~Gaussian noise with zero mean and a covariance matrix $\mat{Q}_{\epsilon}  \negmedspace\in\negmedspace \Real^{N_c\times N_c}$. 


Under this representation, it is widely accepted that surrogates of a wide range of cognitive processes can be decoded from the power of narrow frequency oscillatory sources in $\mat{S}$~\cite{horschig:2014, SPoCdaehne:2014}. We will represent such a surrogate with the row vector $\ve{s}_{z}^\T \negmedspace\in\negmedspace \Real^{N_t}$ of $\mat{S}$, whereas its envelope --- representing power --- will be denoted $\ve{z} \negmedspace\in\negmedspace \Real^{N_t}$ and termed \textit{target variable} as it corresponds to the variable that is to be decoded.

\subsection{Post-hoc labeling of paradigm-agnostic EEG recordings}

We propose a framework that refrains from making (potentially problematic) assumptions about the dynamics of neural activity or the signal-to-noise ratio between an oscillatory source of interest and background sources. The framework relies upon an unsupervised projection of an arbitrary M/EEG dataset $\mat{X}$ onto a source space $\mathcal{S}$ by means of a function $\ve{f}$: 

\begin{align}\label{eq:objfun3}
\begin{split}
  \ve{f}: \mathcal{C} \rightarrow& \,\mathcal{S}\\
 \mat{X} \mapsto& \, \mat{S}.
\end{split}
\end{align}

Assuming we can find such a function which decomposes the M/EEG signals into \textit{reasonable} sources (the next paragraphs will deal with this), we furthermore propose that any source in $\mathcal{S}$ could be selected to serve as the target source $\mat{s}_z$ and that the oscillatory power of this source can be used as labels $\ve{z}$ for the purpose of benchmarking algorithms. While selecting an arbitrary source may sound strange at first, it should be observed that any randomly selected source contained in $\mat{S}$ will have realistic temporal dynamics. Its relative strength, however, will of course vary from source to source.


\subsubsection*{Determining  $\ve{f}$ as a projecting function} 
We propose two strategies for how the function $\ve{f}$ can be selected: the first strategy makes use of an anatomically constrained source space $\mathcal{S}_a \subseteq  \mathcal{S}$ while the second one defines the source space $\mathcal{S}_d \subseteq \mathcal{S}$ in a purely data-driven manner.

\paragraph{Anatomically constrained source space $\mathcal{S}_a$:}
If an anatomically motivated head-model $\mat{A}$, potentially containing a very large number of sources, is available~\cite{hallez2007review}, $\ve{f}$ can be selected such that $\mat{X}$ is projected onto an anatomically constrained version of the source space, $\mathcal{S}_a$. To this end a source reconstruction method may be used. Specifically, the maximum a-posteriori estimate of $\mat{S} \in \mathcal{S}_a$  can be found as the minimizer of the following cost function~\cite{grech2008review, CasHoeMarAnCasHau15}:
\begin{align}\label{eq:objfun_mapping}
	\underset{\mat{S}}{\operatorname{argmin}}\{ || \mat{X} - \mat{A}\mat{S}||_{\mat{Q}}^{2} + \lambda\varTheta\left(\mat{S}\right)\} \;.
\end{align}
Here, $\|\mat{\cdot}\|_{\mat{Q}}$ is the matrix norm of the argument wrt.~$\mat{Q}$, $\lambda \in \Real^{+}$ is a regularization constant, and $\varTheta\left(\mat{S}\right): \mat{S} \mapsto \Real^{+}$ is a penalty term which formalizes the constraints that are imposed upon the neural source activity. Many different algorithms, each with specific choices for $\varTheta\left(\mat{S}\right)$ and $\mat{Q}$, have been introduced~\cite{grech2008review}, each of them representing different priors about the expected characteristics of sources. For the sake of simplicity in the subsequent analysis, we have chosen $\varTheta\left(\mat{S}\right) = ||\mat{S}||_2^2 $ and $\mat{Q} = \mat{I}_{N_s}$, where $\mat{I}_{N_s} \in \Real^{N_s\times N_s}$ is an identity matrix. This approach is commonly termed $\ell_2$-norm regularization~\cite{ng2004feature}, also known as minimum norm estimate (MNE)~\cite{pascual1999review, grech2008review}. Please note that the proposed benchmarking framework is not limited to using the MNE estimate. For this choice, however, it can be shown that the optimal solution for expression \ref{eq:objfun_mapping} is given conveniently by
\begin{align}
\mat{S} = \mat{Q}\mat{A}^\T(\mat{I}_{N_c} + \mat{A}\mat{Q}\mat{A}^\T)^{-1}\mat{X}
\end{align}

\paragraph{Data-driven source space $\mathcal{S}_d$:}
A set of underlying target sources can also be estimated directly from $\mat{X}$ using standard unsupervised linear decomposition methods such as PCA, ICA, factor analysis, among others. It is noteworthy that in these decomposition methods typically the number of sensor channels determines the number of obtainable sources. In the following, we use the fastICA algorithm\cite{hyvarinen2000independent}, which --- among the different blind source separation methods --- has been widely employed for the analysis of neural data~\cite{makeig1996independent, delorme2004eeglab, vigario2000independent}. For this choice, the function $\ve{f}$  is defined as $\mat{S} = \ve{f}(\mat{X}) = \mat{\Phi}\mat{X}$ , where $\mat{\Phi} \in \Real^{N_s \times N_c} $ is a matrix spanning $\mathcal{S}_d$, a space of maximally independent components, where independence is achieved by maximizing non-Gaussianity of the sources $\mat{S} \in \mathcal{S}_d$. Please note again, that the proposed dataset generation framework is not dependent on this specific choice of fastICA.

\subsubsection*{Extraction of target variable $\ve{z}$} \label{subsec:z_extraction}
\begin{figure}[h]
\centering
\includegraphics[width=0.8\textwidth]{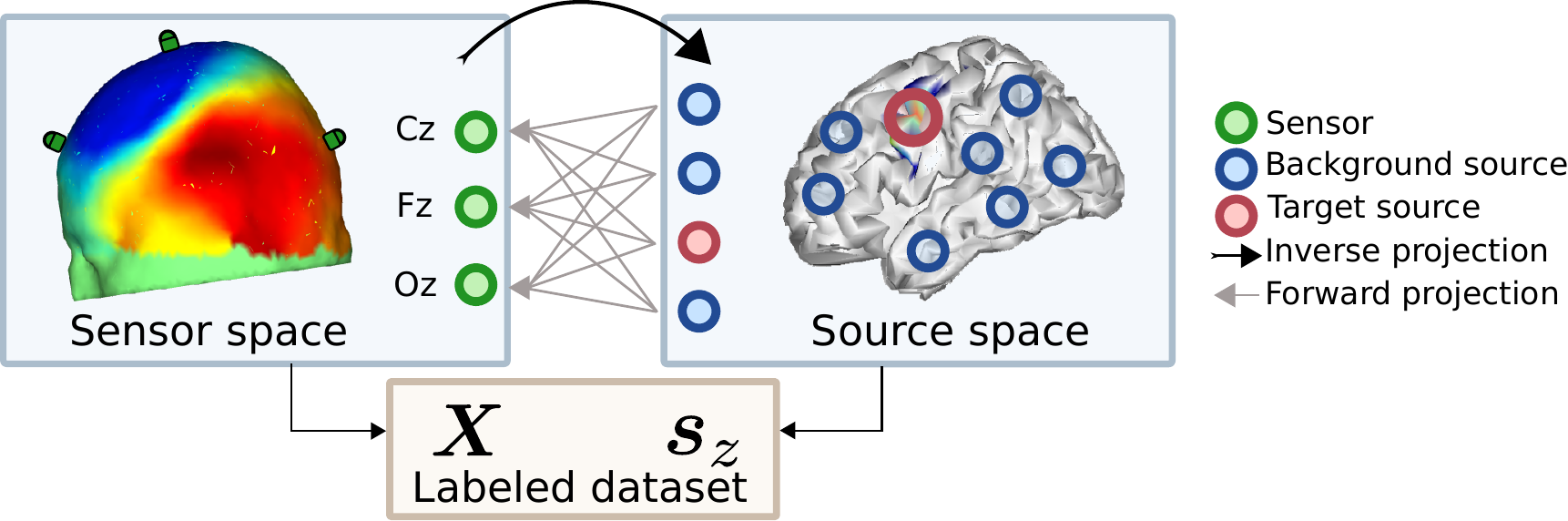}
\caption{Labeled dataset generation framework: Left is the EEG recording represented by a scalp map from a real head-model. The information contained in these signals may be mapped onto the source space. This so-called inverse mapping is the key ingredient for our labeled dataset generation: it is performed either via source reconstruction techniques or by unsupervised decomposition methods.}
\label{fig:genModel}
\end{figure}

Once a set of sources has been determined by either one of the two approaches mentioned above, a target source $\mat{s}_z$ can be selected based upon some prior about the benchmarking problem (e.g.,~specifically strong components only, or components that stem from a brain region known to be involved in the experimental task) or simply by random selection. Then the labels $\ve{z}$ are computed in the following three-step procedure:

\begin{enumerate}
\item Since it is expected that the neural source of interest provides surrogate information of a cognitive process by means of its power in a narrow frequency band~\cite{horschig:2014}, $\mat{X}$ is filtered with a bandpass filter to reflect this assumption and then projected onto $\mathcal{S}$, where a target source  $\ve{s}_z$ is selected.\footnote{Depending on the number of sources to be characterized, the order in which the bandpass filtering and the linear projection is performed can be inverted to improve runtime.}
\item The envelope of the selected source is determined by computing the magnitude of its Hilbert transform 
\begin{align}
\ve{z} = |\mathscr{H} \lbrace \ve{s}_z \rbrace|
\end{align}
\item If desired, the data and the labels can be windowed into epochs. It is important to remark that this step depends on the algorithm which will be benchmarked. 
\end{enumerate}
In this way, we obtain a dataset consisting of real EEG recordings $\mat{X}$ and a noiseless variable $\ve{z}$ containing the target labels. Using $\mat{X}$ and $\ve{z}$, any arbitrary supervised decoding algorithm can be developed, benchmarked and validated. Figure~\ref{fig:genModel} illustrates the general idea of the proposed labeled dataset generation framework.


\section{Experimental Setup}
As an illustrative example for a typical use scenario of our dataset generation framework, we will benchmark the source-power comodulation (SPoC) algorithm\cite{SPoCdaehne:2014} wrt.~several parameters of the simulation, i.e.,~dataset size, label noise level, and relative power of the target source. In the following section we will describe the generation of the labeled dataset as well as details about the  SPoC implementation.

\subsection{The EEG dataset}
\paragraph{Signal acquisition}
The EEG signals used in the present use-case were recorded from seven healthy subjects (three females) with a mean age of 28 years. Seventy three minutes of EEG data on average were recorded while subjects sat calmly in front of a computer screen and performed the sequential visual isometric pinch task (SVIPT)~\cite{reis2009,MeiCasReiTan16}. Given the paradigm-agnostic character of the post-hoc labeling framework, details about the paradigm remain outside the scope of this paper. EEG signals were recorded from 31 passive Ag/AgCl electrodes (EasyCap GmbH, Germany) placed according to the extended 10-20 system. Impedances were kept below 20\,k$\Omega$. All channels were referenced against the nose during recording and were re-referenced against the EEG common average during the post-hoc analysis. The EEG signals were registered by BrainAmp DC amplifiers (Brain Products GmbH, Germany) at a sampling rate of 1\,kHz, with an analog lowpass filter of $250\,$Hz applied before digitization.

\paragraph{Pre-processing specifically for the labeled dataset generation} All the EEG-processing steps were performed using the BBCI toolbox~\cite{blankertz2016berlin}. EEG Signals were bandpass filtered between 0.2 and 48\,Hz with a 5-th order Butterworth filter and then sub-sampled at 120\,Hz. Assuming the alpha band being in the focus of a benchmarking scenario, EEG data were further bandpass filtered  with a 5-th order Butterworth filter to this target frequency of 8 to 12\,Hz. This target frequency band can be modified according to different analysis goals, but in our case, for the sake of compactness in the use-case analysis, we have kept its value fixed. Finally, the procedure described in section \ref{subsec:z_extraction} was applied in order to obtain a labeled dataset comprised by $\mat{X}$ and $\ve{z}$, both for the head-model based and the data-driven dataset generation framework.

For the generation of the datasets containing anatomical constraints on the sources, the publicly available \textit{New York Head}~\cite{HuangParraHaufe15} was used. It describes a finite element model containing 2000 sources located on the cortical surface. The sources are subsampled from a highly detailed model containing 74382 sources, which had been computed from a non-linear average of 152 human brains. The New York Head-model takes scalp, skull, cerebro-spinal fluid, gray matter, and white matter into account. Sources were assumed to be perpendicularly oriented wrt.~the cortical surface, however, our framework could be also used with models that allow for free source orientation.

On the other hand, for the data-driven approach, a fixed number of 20 ICA components were extracted. Afterwards, only components corresponding to actual neural sources were selected for further analysis. For the identification of such neural components, the multiple artifact rejection algorithm (MARA)~\cite{winkler2011automatic} was applied.

\paragraph{Pre-processing of data for SPoC} For outlier detection, the continuous EEG data $\mat{X}$ were bandpass filtered between $0.7$ and $25\,$Hz with a 5-th order Butterworth filter. As SPoC is sensitive for outliers, segments of the continuous data with peak-to-peak amplitude $> 80\,\mu$V were marked as artifactual for later removal in the processing pipeline. 

Starting once more with the original continuous data, a bandpass filter around the target frequency (between $8$ and $12\,$Hz) with a 5-th order Butterworth filter was applied. Then, EEG data and the target source $\ve{z}$ were segmented in non-overlapping windows of $1\,s$ duration. At this point epochs marked earlier as artefactual were removed. For the remaining segments, the epoch-wise average power of $\ve{z}$ was extracted and used as the target variable to train SPoC.

At this point we need to remark, that the epoching does not necessarily need to obey the original time structure of the experimental paradigm, under which $\mat{X}$ was recorded. Instead, epochs could be defined freely in order to serve the goals of the benchmarking.

\subsection{ The decoding method: Source-Power Comodulation - SPoC}
The multivariate neural decoding method called source power comodulation (SPoC)~\cite{SPoCdaehne:2014} utilizes a supervised regression approach in order to estimate a set of spatial filters $\mat{W} \in \Real^{N_c\times N_f}$ , onto which $\mat{X}$ will be linearly projected to extract the underlying target source(s) and decode the target variable(s) in the form of source power.



Assuming that M/EEG data $\mat{X}$ have been bandpass filtered to the frequency band of interest and segmented into a set of $N$ epochs, where $\mat{X}(e)$ and $\ve{z}(e)$ represent the $e$-th epoch of the M/EEG data and the corresponding target variable (label), then a spatial filter $\mat{w} \in \Real^{N_c}$ (which is a column vector in $\mat{W}$) is optimized such that the power of an epoch $\Theta_x(e) = \operatorname{var}[\hat{\ve{s}}](e)$ of the spatially filtered data $\hat{\ve{s}}(e)=\mat{w}^\top\mat{X}(e)$,  maximally covaries with the target variable $\ve{z}$:

\begin{align}\label{eq:opti_prob}
\argmax_\ve{w} \lbrace \mathrm{cov} \left[ \Theta_x(e), \ve{z}(e) \right] \rbrace \, \forall \, e
\end{align}
 Based on data of multiple epochs, it can be shown that solving this optimization problem is equivalent to solving the generalized eigenvalue problem
\begin{align}\label{eq:eig_spoc}
\mat{C}_z\mat{W} = \mat{\Lambda}\mat{C}\mat{W}
\end{align}
where $\mat{C} = N^{-1}\sum_e^N\mat{X}(e)\mat{X}(e)^\T$ is the covariance matrix of $\mat{X}$ and $\mat{C}_z = N^{-1} \sum_e^N  z(e)\mat{X}(e)\mat{X}(e)^\T$ the epoch-wise $\ve{z}$-weighted covariance matrix. Likewise, $\mat{\Lambda} \in \Real^{N_c\times N_c}$ is a diagonal matrix containing  in its main diagonal the eigenvalues of the formulated problem.

Given a spatial filter $\mat{w}_{tr}$ determined on training data, the true target variable $\ve{z}$ can subsequently be estimated as $\hat{\ve{z}}$ on a single-epoch basis for unseen test data $\mat{X}_{te}$ via $\hat{\ve{z}}(e) = \operatorname{var}[\mat{w}_{tr}^\top\mat{X}_{te}(e)]$. 

The results reported here were computed using the strongest SPoC component only, i.e.,~$\ve{w}$ given by the eigenvector related to the largest eigenvalue obtained by solving  Eq.~\ref{eq:eig_spoc}. Components ranked lower in terms of their eigenvalues were discarded. 
Furthermore, since the SPoC algorithm is a linear method and follows the generative model of the EEG shown in Eq.~\ref{eq:genModel}, it is possible to interpret the spatial pattern of a computed SPoC component. The pattern can be estimated via $ \hat{\ve{a}}_z = \mat{C} \ve{w} \left( \ve{w}^\top \mat{C} \ve{w} \right)^{-1}$~\cite{SPoCdaehne:2014,haufe2014interpretation}. Later on, we will refer to this estimated spatial pattern $\hat{\ve{a}}_z$ to assess the spatial accuracy of SPoC.

\paragraph{Decoding accuracy}
The accuracy of the decoding estimation provided by SPoC is assessed in terms of two measures: First, the estimated and the true band power of the target source can be compared. This is realized by calculating the linear correlation $\rho$ between the estimated target variable $\est{\ve{z}}$ and the known true labels $\ve{z}$. Second, since the method allows for the interpretation of the model weights $\ve{w}$ as spatial patterns, the angle $\alpha$ between the estimated spatial pattern $\hat{\ve{a}}_z $ and its ground truth $\ve{a}_{z}$ is a useful performance measure. It is defined as
\begin{align}
\alpha_r = &\arccos \left( \frac{\ve{a}_{z}^\T\hat{\ve{a}}_z}{||\ve{a}_z||\,||\hat{\ve{a}}_z||} \right)  \nonumber  \\
\alpha = &
\begin{cases} 
     \alpha_r, & \alpha_r \leq \pi/2 \\
      \pi - \alpha_r, & \alpha_r > \pi/2 \\
   \end{cases}
\end{align}

Note that both performance measures are specific for SPoC and should be replaced in other benchmarking scenarios depending on the utilized decoding task and algorithm. For example, mean-squared error, classification accuracy (if a classification problem is at hand), earth mover's distance for spatial assessment of the patterns, among others, are examples of measures that may alternatively describe the decoding accuracy of the algorithm studied.

\subsection{Hyperparameter sweep}
To exemplify a typical use-case for the labeled dataset generation framework, we tested the robustness of SPoC wrt.~different hyperparameters of the framework which influence the decoding performance. Sensitivity wrt.~the number of training epochs available was evaluated by sweeping from 50 to 2000 epochs. The influence of label noise was investigated by varying $\xi_n  \in \Real$ in the range $0 < \xi_n < 1$ as a variable determining the amount of label noise, which is to to be added to $\ve{z}$.  The variable $\xi_n$ controls the correlation $\rho_n$ between the original clean labels $\ve{z}$ and labels $\ve{z}_n$ which have been contaminated with a certain amount of noise, such that $\xi_n = (1-\rho_n)$. Subsequently, noisy labels $\ve{z}_n$ are defined as
\begin{align}
\ve{z}_{n} = \ve{z} + \frac{1- (1-\xi_n)^2}{(1-\xi_n)^2} \mathrm{var}\left( \ve{z}\right) \ve{\eta}
\end{align} 
where $\ve{\eta}$ is a normally distributed random variable.

Manipulating label noise and the amount of data samples allows to challenge a decoding algorithm concerning two problems commonly found in the field of neural decoding: 1) only a limited amount of suitable training data is available due to inherent constraints during data acquisition and 2) a lack of ground truth for training labels, i.e.,~the target variable is typically contaminated with confounds and does not perfectly correlate with a specific neural source.

Decoding algorithms can also be studied by considering the characteristics of the target sources. As an example, we have analyzed the performance of the decoding algorithm SPoC wrt.~the strength of the target source as explained by its relative power, which could be seen as another hyperparameter of the dataset generation framework. Sources can be characterized in the range between 0 representing the weakest source among all sources and 1 as the strongest one. We term this parameter \textit{relative source power}. It is worth noting that the ranking of the sources provided by the relative power highly correlates with the variability of the power of the sources themselves.

These kind of hyperparameters, i.e.,~label noise level, dataset size and relative source power, are typically not controllable in a real-case application, in which the best possible decoding performance is desired. However, in a benchmarking scenario, absolute control of such hyperparameters is important since it allows the characterization of a decoding algorithm in such a way that it is possible to determine under which conditions its use is still suitable.

The influence of the hyperparameters of the framework upon decoding performance was estimated using functional ANOVA\cite{HutHooLey14}. The sweep over label noise, dataset size and relative source power was performed using the random online adaptive racing hyperparameter optimization procedure~\cite{hutter2011sequential}, obtaining 1300 evaluations of different hyperparameter configurations. Each evaluation corresponds to the average performance obtained in a chronological 5-fold cross-validation procedure. All results shown were marginalized across the 7 subjects contained in the dataset.

\section{Results of use-case: SPoC}
\def \sizepowerdist {0.3}
\begin{figure}
\centering
\begin{tabular}{c|cc}
 Artifact rejection log  & \multicolumn{2}{c}{Relative source power distribution} \\ 
& Head-model& Data-driven  \\

\includegraphics[width=\sizepowerdist\textwidth]{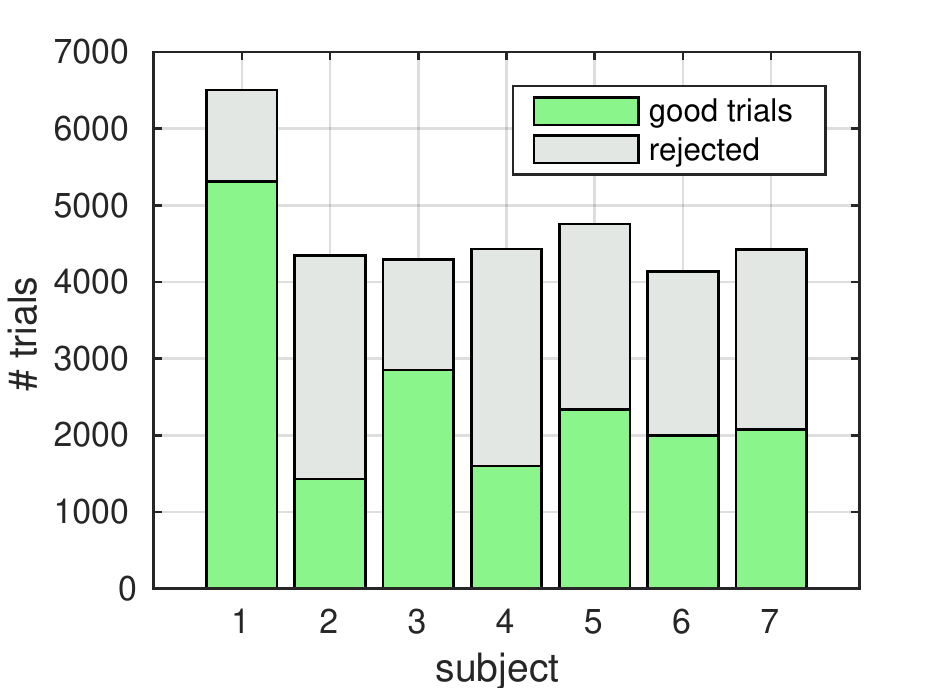} &
\includegraphics[width=\sizepowerdist\textwidth]{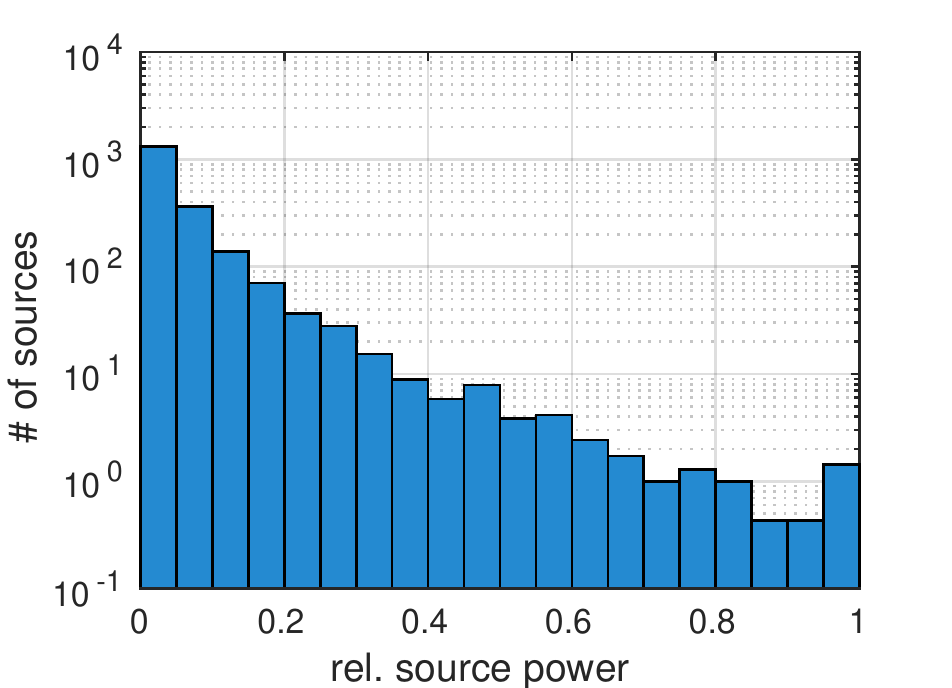} &
\includegraphics[width=\sizepowerdist\textwidth]{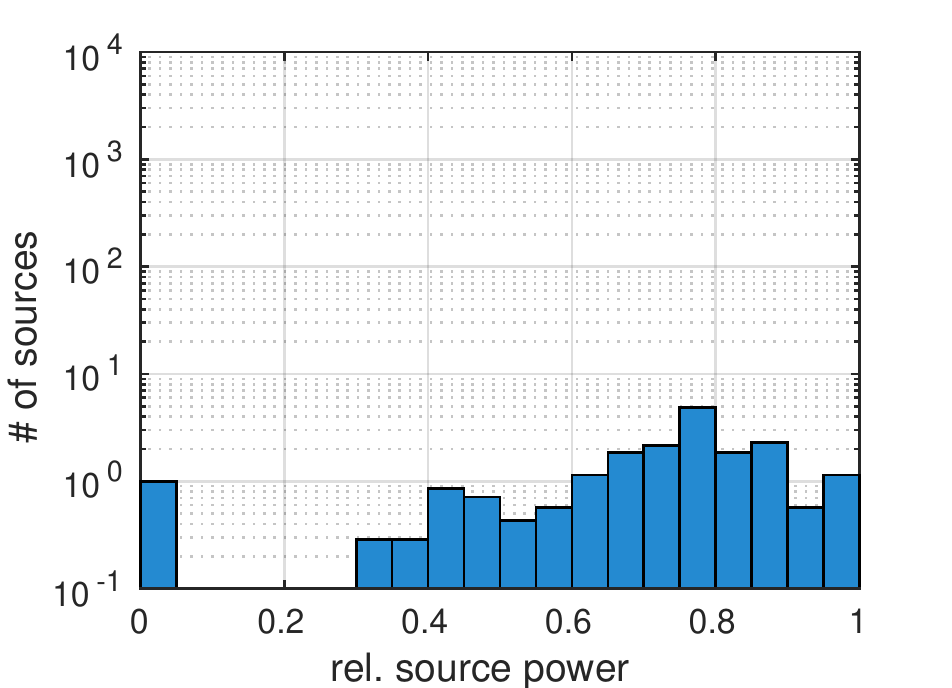}
\end{tabular}
\caption{Characteristics of the generated datasets. Left: Good and artifactual epochs for each of the subjects contained in the original EEG dataset. Right: Distribution of the relative source power across all sources for head-model based and data-driven based generation (average across the seven subjects).}
\label{fig:stats_dataset}
\end{figure}

Figure \ref{fig:stats_dataset} shows an overview of the datasets generated using our framework. On average, 4700 epochs were obtained from only $73$\,min of EEG recordings, from which 2136 epochs were marked as artifactual per subject, yielding a rejection rate of approximately $45\,\%$. Furthermore, the power distribution of the sources varies greatly between the head-model and the data-driven based generators. In the head-model, the amount of weakly activated sources dominates, whereas the power distribution is more uniform in the data-driven approach.

Figure \ref{fig:res_hd_par_sweep} illustrates the results obtained for the hyperparameter sweep performed for the labeled dataset generator with anatomical constraints on the sources, whereas Figure \ref{fig:res_dd_par_sweep} shows the analogue results obtained for the data-driven based generator. The results exemplify two use-cases for our labeled dataset generation framework, where the properties of a given decoding algorithm can be analyzed under different experimental conditions. Despite of the difference in relative source power, the observed results were similar for both generator types (head-model based and data-driven based). It can be seen that SPoC's performance is highly sensitive to the relative power of the target source and, as expected, a higher decoding accuracy was achieved for stronger sources. In terms of training data samples, a correlation between performance and dataset size can be seen for datasets containing less than $\approx$ 750 samples, whereas having datasets with more samples was not beneficial for achieving even higher decoding accuracy. Finally, it is also interesting to see how SPoC behaves under varying label noise. We observed a floor effect for $\xi_n > 0.9$ mainly in the data-driven based dataset generator. For smaller values of $\xi_n$, $\rho$ performance improves linearly as $\xi_n$ decreases. This effect is not clearly observable in the spatial accuracy measure $\alpha$.

\newcommand{\rulesep}{\unskip\ \vrule height -1ex\ }
\def \size {0.45}
\begin{figure}[h!]
\centering
\begin{tabular}{c|c}
Performance $\rho$ & Performance $\alpha$\\
\includegraphics[width=\size\textwidth]{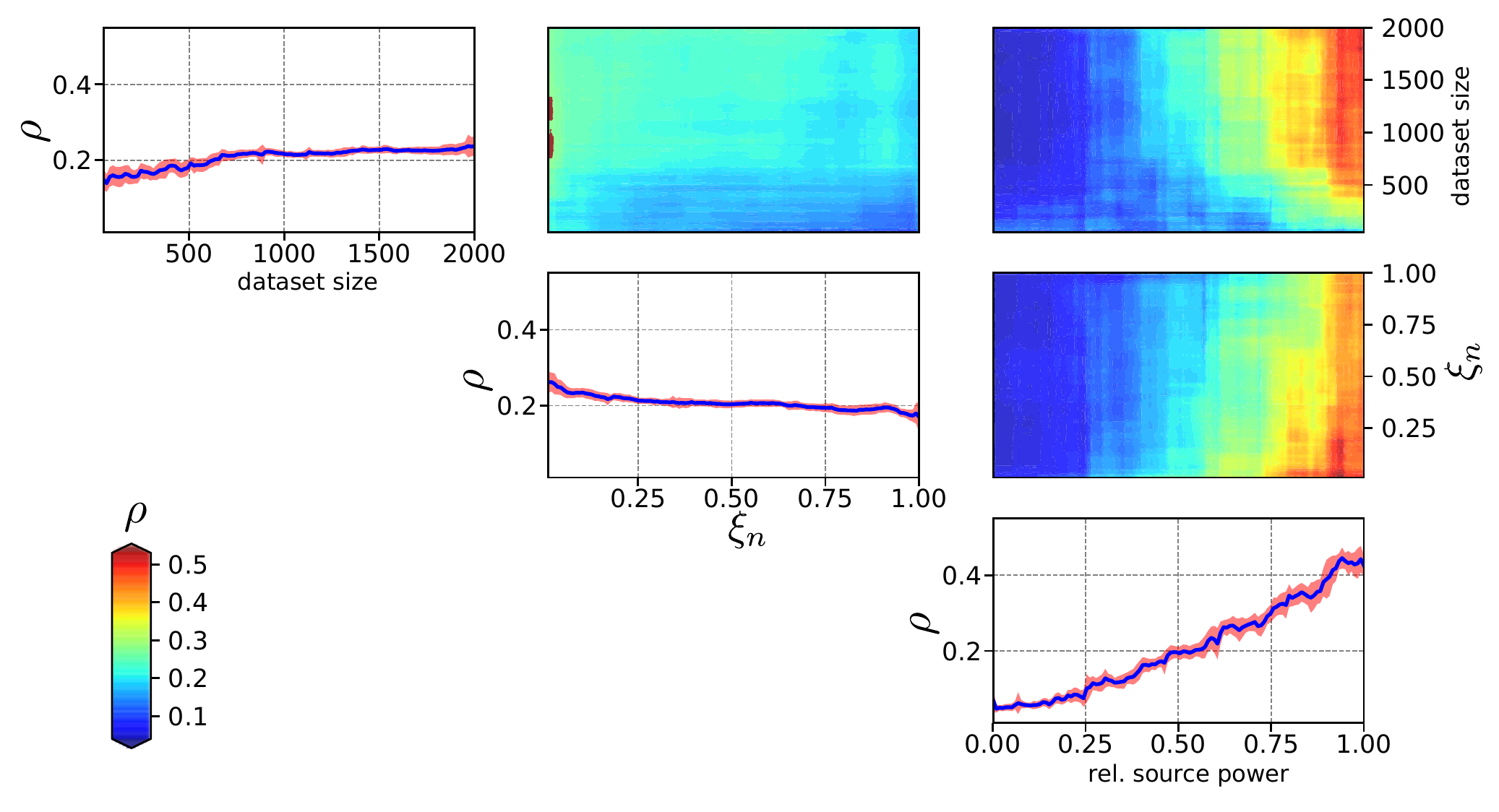} &
\includegraphics[width=\size\textwidth]{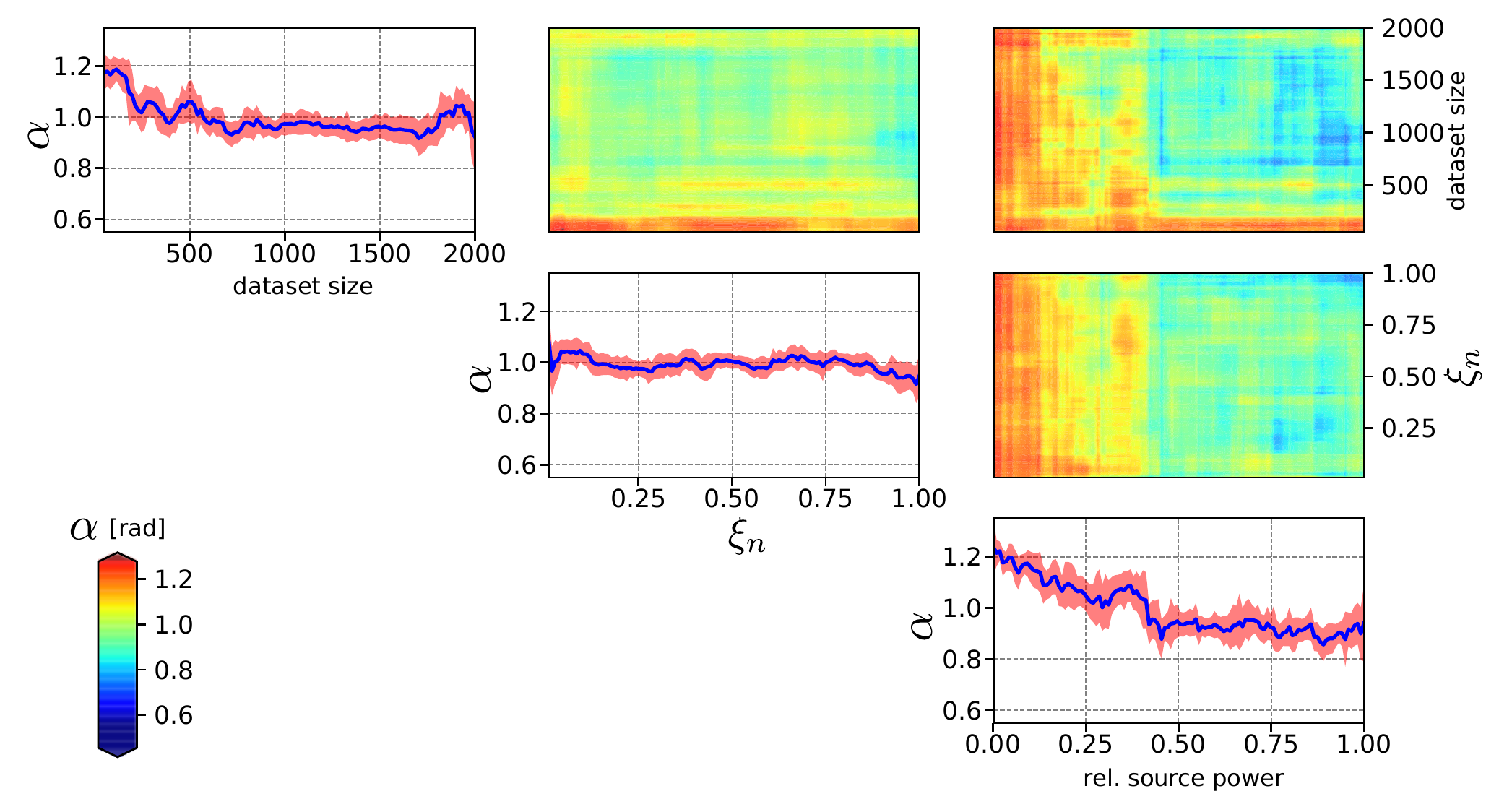}
\end{tabular}
\caption{Head-model based simulation: Sensitivity of SPoC to hyperparameters. Left: evaluated according to correlation metric $\rho$; Right: evaluation according to angle $\alpha$ between patterns. In every main diagonal, the marginalized performance for dataset size, label noise level $\xi_n$ and relative power of target source is depicted. Off-diagonal plots visualize the corresponding pair-wise marginalized performances.}
\label{fig:res_hd_par_sweep}
\end{figure}

\begin{figure}[h!]
\centering
\begin{tabular}{c|c}
Performance $\rho$ & Performance $\alpha$ \\

\includegraphics[width=\size\textwidth]{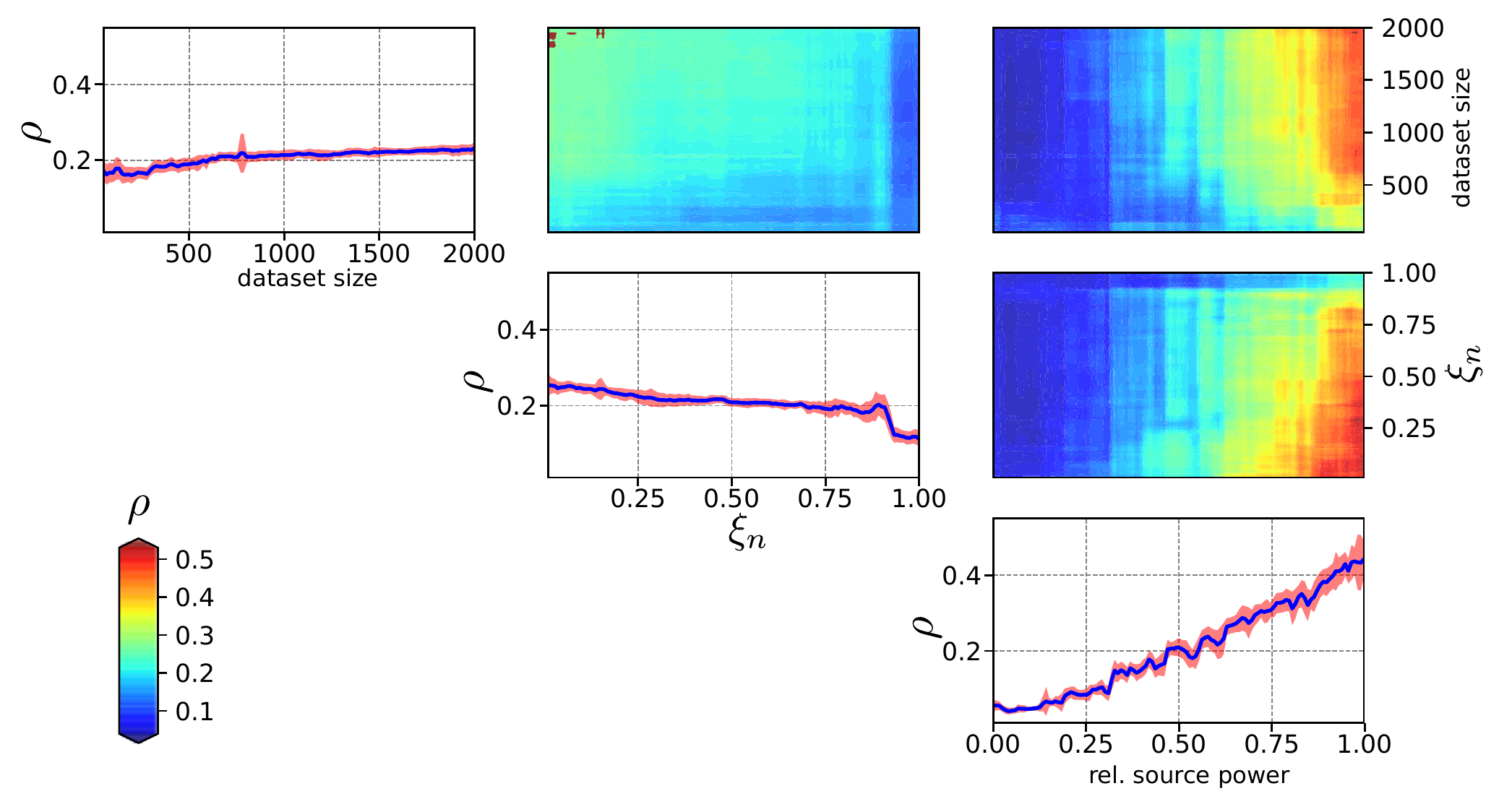}&
\includegraphics[width=\size\textwidth]{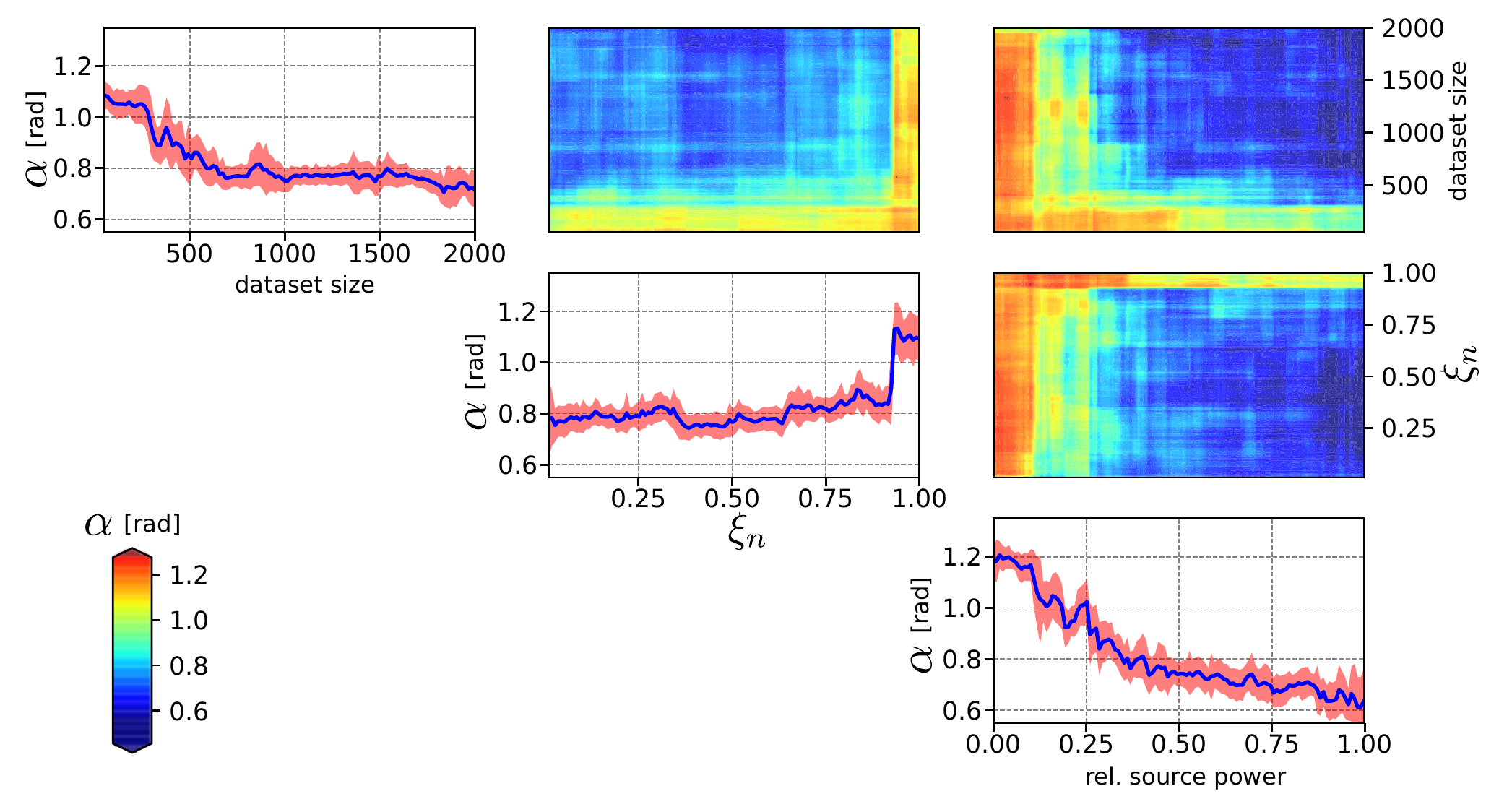}
\end{tabular}
\caption{
Data-driven based simulation: Sensitivity of SPoC to hyperparameters. For a description, please see Figure~\ref{fig:res_hd_par_sweep}.}
\label{fig:res_dd_par_sweep}
\end{figure}

\section{Discussion}
Methods for decoding oscillatory neural sources from M/EEG signals are valuable tools that allow to expand the understanding about underlying neural processes, which may have corresponding behavioral, sensorial, and motor surrogates. To develop these methods it is necessary to have a suitable testbench that allows to run objective benchmarks in a realistic scenario. An ideal testbench should provide a large amount of data, clean labels and realistic neural dynamics. Unfortunately, state-of-the-art approaches lack one or several of these properties: On the one hand, synthetically generated data could in principle provide infinitely large datasets and noiseless levels, but have the drawback that strong assumptions about neural dynamics need to be made. On the other hand, using real M/EEG recordings gives access to real neural dynamics but comes at the cost of noisy labels and small amounts of data.

To solve the aforementioned issues, we have introduced a labeled dataset generation framework for benchmarking. It implements a  post-hoc labeling of (potentially very long) paradigm-agnostic pre-recorded M/EEG signals.
The framework allows to generate labeled datasets containing real neural signals and by doing so, it prescinds from making critical assumptions about neural dynamics.
As labels provided by the framework are extracted directly from the data, they are noise free, i.e.,~they describe the neural target source perfectly.
Furthermore, the post-hoc labeling of paradigm-agnostic M/EEG recordings offers greater efficiency in terms of data usage.  Compared to real datasets whose labels depend on the paradigm they were recorded on, our post-hoc labeling can make use also of idle periods or preparatory intervals. For the example we made, this led to an exploitation of effectively 55\% of the overall M/EEG recording time for dataset generation.

The proposed framework allows to generate datasets based on arbitrary subspace projections. To support the adoption of our framework by practitioners, we exemplified two different projection strategies for the extraction of labels: a data-driven and an anatomically motivated strategy.

Our framework allows absolute control over the important hyperparameter label noise level of the generated dataset and provides full knowledge about the statistical and --- in case of the head-model ---anatomical properties of the target sources.
It provides an ideal framework for comparing competing decoding methods, as it yields insight into the data conditions under which each method stands out among the others. Examples are the required (minimum) amount of training data or feasible levels of label noise. Specific expectations about target sources (prior knowlege about central frequency, strength, or anatomical location etc.) can be incorporated in the generation of the benchmark datasets to fine-tune the search for the best decoding method.
Last but not least, knowing the limitations of a decoding method may pave the road to its improvement.

However, one relevant parameter unfortunately remains outside the control of our framework: the amount of sensor noise is determined by the available real-M/EEG signals and can not be improved (only worsened) post-hoc. 
Here, synthetic data generation approaches have a theoretical advantage, as they can control the level of sensor noise. In practice, however, it may not be straightforward to determine a noise levels during synthetic data generation in order to match real experimental conditions.

Table \ref{tab:sim_comparison} visually summarizes the properties of our contribution (\textit{post-hoc labeled data}) compared to other testbench approaches.

\newcommand{\tikzcircle}[2][red,fill=red]{\tikz[baseline=-0.7ex]\draw[#1,radius=#2] (0,0) circle ;}

\begin{table}[h]
\centering
\begin{tabular}{|c|c|c|c|c|}
\hline
 & Dataset size & Label noise ctrl.   & Sensor noise ctrl.  & Real statistics\\ 
 \hline
Synthetic data & Large \tikzcircle[black,fill=green]{2pt} & Yes \tikzcircle[black,fill=green]{2pt} & Yes \tikzcircle[black,fill=green]{2pt} & No \tikzcircle[black,fill=red]{2pt} \\ 
Real EEG data & Small \tikzcircle[black,fill=red]{2pt} & No \tikzcircle[black,fill=red]{2pt} & No \tikzcircle[black,fill=red]{2pt} & Yes \tikzcircle[black,fill=green]{2pt} \\ 
\textit{Post-hoc labeled data} & Medium/large \tikzcircle[black,fill=yellow]{2pt}/\tikzcircle[black,fill=green]{2pt} & Yes \tikzcircle[black,fill=green]{2pt}  & No \tikzcircle[black,fill=red]{2pt} & Yes \tikzcircle[black,fill=green]{2pt} \\
\hline
\end{tabular} 
\caption{Comparison of advantages and disadvantages of two state-of-the-art testbench scenarios against the proposed novel \textit{post-hoc labeling} framework.}
\label{tab:sim_comparison}
\end{table}

We envision the adoption of the proposed framework as a tool for development and testing of decoding algorithms for oscillatory neural phenomena. For practitioners, we provide the source code and EEG recordings utilized in our example for download\footnote{https://github.com/bsdlab/Post-HocLabeling}.

\section{Acknowledgements}
We would specially like to thank Frank Hutter and Katharina Eggensperger for providing the tools for the hyperparameter sweep and functional ANOVA.

This work was supported by BrainLinks-BrainTools Cluster of Excellence funded by the German Research Foundation (DFG, grant number EXC 1086). The authors also acknowledge support by the state of Baden-W\"urttemberg, Germany, through bwHPC and the German Research Foundation (DFG) through grant INST 39/963-1 FUGG. 

\bibliographystyle{unsrt} %
\bibliography{bsdlab_general,general}
\end{document}